# Persian Handwritten Digit, Character and Word Recognition Using Deep Learning

**Mehdi Bonyani[a], Simindokht Jahangard[b*], Morteza Daneshmand[c]**
[a] Department of Computer Engineering, University of Tabriz, Iran, m_bonyani96@ms.tabrizu.ac.ir
[b] Department of Robotics Engineering, Amirkabir University of Technology, Iran, s_jahangard@aut.ac.ir
[c] Institute of Technology, University of Tartu, Estonia, mortezad@ut.ee

## Abstract

Digit, letter and word recognition for a particular script has various applications in today's commercial contexts. Nevertheless, only a limited number of relevant studies have dealt with Persian scripts. In this paper, deep neural networks are utilized through various DensNet architectures, as well as the Xception, are adopted, modified and further boosted through data augmentation and test time augmentation, in order to come up with an optical character recognition accounting for the particularities of the Persian language and the corresponding handwritings. Taking advantage of dividing the databases to training, validation and test sets, as well as k-fold cross validation, the comparison of the proposed method with various state-of-the-art alternatives is performed on the basis of the HODA and Sadri databases, which offer the most comprehensive collection of samples in terms of the various handwriting styles possessed by different human beings, as well as different forms each letter may take, which depend on its position within a word. On the HODA database, we achieve recognition rates of 99.72% and 89.99% for digits and characters, being 99.72%, 98.32% and 98.82% for digits, characters and words from the Sadri database, respectively.

## 1) Introduction

Since the early 1970s, the recognition of written patterns has attracted many researchers. Although this has led to the development and expansion of numerous efficient algorithms for e.g. typesetting purposes, most of them are still not fully reliable, and require further enhancements [1]. Optical Character Recognition (OCR) is the process of converting scanned images of handwritten or machine-printed text to sequences of characters that can be read by the machine, or stored in simple- or hyper-text formats, e.g. in markup languages. A typical OCR pipeline consists of three main modules, namely, document digitization, character or word recognition, and output distribution [2].

When it comes to Persian and Arabic languages, which constitute the main focus of this paper, in spite of the fact that they are deemed to be verbally and grammatically different, they share most of their characters and digits. They are widely utilized in numerous countries, especially in the Middle East. For the majority of the inhabitants of countries such as Iran, Tajikistan and Afghanistan, i.e. more than 667 million people [3], Persian is the mother tongue, while Arabic is the dominant language in countries such as Saudi Arabia, Yemen, Iraq and Oman.

There are various factors which exacerbate the complications associated with the recognition of pieces of Persian or Arabic text. First and foremost, each digit may be written in several different ways. This concept has been illustrated in Fig. 1(a). Moreover, in writings associated with the foregoing languages, most of the letters constituting a word are deformed from the original shape and stitched to each other. Consequently, the shape of each letter may depend on its location within the relevant word, i.e. on whether it is the first letter in the word, the last one, or any of the rest. The latter notion has been illustrated in Fig. 1(b) and 1(b). Additionally, there are various styles of handwriting, which further complicates the task of recognizing the characters or words [4].

Moreover, the nuances pertaining to the certain calligraphy utilized by the person at hand compounds the troubling factors associated with feature extraction. Furthermore, OCR methods that could operate reliably through hierarchical learning, especially in the presence of noise, are quite rare, and on top of that, suitable databases that could be utilized for developing and testing such algorithms are not abundant.

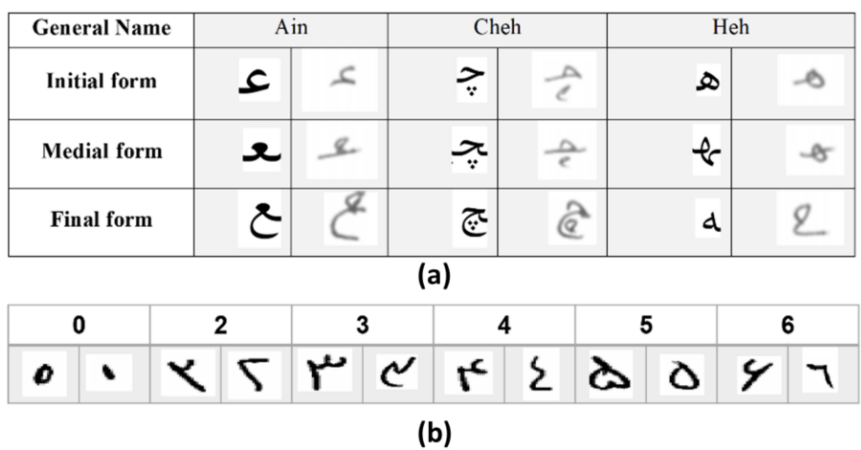

Figure 1. Challenges in the recognition of handwritten Persian characters: (a) Different styles of writing some digits, (b) Different appearances of some letters, depending on their location within a word.

Handwriting recognition may be performed either offline or online. The former concerns studying a piece of text that has been written already, which will then be scanned and processed, while the latter takes place on the go, e.g. where a stylus and touch screen are utilized [1]. In this paper, some of the aforementioned problems are alleviated for the sake of coming up with a relatively accurate system for recognizing handwritten Persian digits, letters and words via deep learning by means of Convolutional Neural Networks (CNNs). The latter rely on automatically determining optimally distinctive low-level features, being subsequently integrated into the learning procedure to determine high-level ones [5], and have lately resulted in significantly superior recognition rates compared to classic approaches in which feature descriptors are devised and supplied explicitly by human beings [6, 7].

To this end, various configurations of Densely Connected Convolutional Networks (DenseNet) [8] architectures, which are obtained by adjusting the number and types of layers, the size of each layer and the number of neurons associated with it and the type of the activation functions, are tried out and compared in terms of the resulting recognition rates. In order to demonstrate its unprecedented performance, the OCR system is finally tested on some of the most comprehensive databases, including HODA and Sadri [9, 10]. They have been selected for the purpose of this study due to the fact that among the existing databases of handwritten Persian scripts, they offer samples with the most diverse range of properties in terms of the aforementioned differences between various handwriting styles.

The rest of the paper is organized as follows. In Section 2, the existing literature and state-of-the-art in OCR are concisely reviewed. In Section 3, the proposed method is described from a technical perspective, including the details of the preprocessing and training procedures, along with the evaluation metrics and the relevant databases. In section 4, the results are presented and evaluated against the ones reported in the literature heretofore. Finally, in Section 5, the conclusions are listed, and suggestions on possible future research directions are put forward.

## 2) Literature review

As aforementioned, classical methods aimed at OCR for handwritten Persian text mostly rely on explicit construction of features, such as geometric and correlation-based ones. Sadri et al. [10] used the gradient histogram method, and applied MultiLayer Perceptron (MLP) neural networks and nearest-neighbors classifiers, resulting in a maximal recognition rate of 98.57% on digits. Parseh et al. combined features including zoning, crossing count and hole size from the HODA database, and utilized Principal Component Analysis (PCA) to reduce the dimensions of the feature space. They employed Support Vector Machines (SVMs) for classification, and achieved a recognition rate of 99.07% [11] on digits. A Holography graph neuron-based system was proposed by Hajihashemi et al [12], where vector symbolic architectures led to a recognition rate of 88.9%, again on the HODA dataset. OCR based on Hidden Markov Models (HMMs) was studied by Ghods and Sohrabi [13], analyzing the main body as the first step, which was then followed by the final recognition through delayed strokes , e.g. dots and small signs, resulting in recognition rates of 95.9% and 94.2% on the groups and characters of the TMU database, respectively. Arani [14] fused the outputs of Right-to-Left (RtL) and Left-to-Right (LtR) of HMMs on a database of the names of cities, accomplishing a recognition rate of 97.93%. Khorashadizadeh et al. [4] combined four directional Chain Code Histograms (CCHs) and Historam of Oriented Gradient (HOG) to come up with an optimal OCR system. Their proposed feature set has 164 dimensions, and applying SVM to the HODA database, a recognition rate of 99.58% was obtained. Sajedi et al. [15] proposed a modified framing feature to recognize digits, and made use of SVM and KNN classifiers, achieving a recognition rate of 99.07% on the HODA database. Montazer et al.[16]  utilized Bag of Visual Words (BoVW) together with Scale Invariant Feature Transform

(SIFT) and Quantum NNs (QNNs) as the classifier, leading to a recognition rate of 99.30% on the digits of the HODA database.

Nevertheless, as mentioned before, in more recent studies, it has been preferred to resort to deep learning through CNNs, in order to automatize the process of coming up with the list of most distinctive features. Parseh et al. integrated a non-linear multi-class SVM classifier instead of a last fully connected layer in their CNN structure, as a result of which the recognition rate was increased to 99.56% on the HODA database [17]. Nanehkaran et al. evaluated the performance of traditional methods such as K-nearest neighbor, Artificial NNs (ANNs) and SVMs in recognizing digits, and realized that CNNs and CNN-based autoencoders perform more efficiently than those, where a recognition rate of 99.3%, which was obtained through SVMs, was increased to 99.45% [18]. Gadikolaie et al. [19] presented a segmentation-based approach for offline recognition of handwritten Persian words, where each word was divided into sub-words, followed by applying a Recurrent NN (RNN) with True/False outputs for each sub-word, instead of using a single complex classifier with multiple output classes. They evaluated their method on the Iranshahr database, with the best recognition rate being 90.5%. Akhlaghi et al. [20] employed NNs and CNNs, where three convolutional layers and two pooling layers were applied to the digits of the HODA database, and achieved a recognition rate of 99.34%. Savaramini et al. [21] obtained a recognition rate of 97.7% on the digits of the same database. Roohi et al [22] considered three types of NNs, namely, a linear softmax classifier, a CNN based on the LeNet-5 structure and an extended CNN that is based on a bagging paradigm, with a recognition rate of 97.1% on the letters of the HODA database. Farahbakhsh et al. used the Alexnet along with data augmentation, and come up with a recognition rate of 99.67% on the digits of the same database [23]. Bossaghzadeh et al. applied the VGG NN to the HODA database subsequent to applying PCA to the preliminary features, aiming at reducing the computational load, followed by linear SVM, resulting in a recognition rate of 99.69% [24]. Table 1 provides a summary of the studies discussed throughout this section.

*Table 1. Summary of the OCR rates reported in the literature on handwritten Persian digits, letters and words.*

| Author | Data type | Database | OCR rate | Method |
|---|---|---|---|---|
| **Sadri et al. [10]** | Digits | Sadri | 98.57% | MLP NN, nearest neighbors |
| **Parseh et al. [11]** | Digits | HODA | 99.07% | PCA, SVM |
| **Ghods et al. [13]** | Letters | TMU | 95.90%, 94.20% | HMM |
| **Parseh et al. [17]** | Digits | HODA | 99.56% | CNN, SVM |
| **Nanehkaran et al. [18]** | Digits | HODA | 99.3% | Geometric and correlation-based features |
| **Nanehkaran et al. [18]** | Digits | HODA | 99.45% | CNN, Auto-encoder |
| **Arani et al. [14]** | Words | Iranshahr | 97.93% | RtL- and LtR-HMM |
| **Ghadikolaie et al. [19]** | Words | Iranshahr | 90.50% | RNN |
| **Khorashadizadeh et al. [4]** | Digits | HODA | 99.58% | CCH, HOG |
| **Sajedi et al. [15]** | Digits | HODA | 99.07% | Modified framing features, SVM, KNN |
| **Akhlaghi et al. [20]** | Digits | HODA | 99.34% | NN, CNN |
| **Montazer et al. [16]** | Digits | HODA | 99.30% | BoVW, SIFT, QNN |
| **Sarvaramini et al. [21]** | Letters | HODA | 97.7% | CNN |
| **Roohi et al. [22]** | Letters | HODA | 97.10% | Linear classification, LeNet-5 CNN |
| **Hajihashemi et al. [12]** | Letters | HODA | 88.90 % | Holography graph neuron-based system |
| **Farahbakhsh et al. [23]** | Digits | HODA | 99.67% | Alexnet |
| **Bossaghzadeh et al. [24]** | Digits | HODA | 99.69% | VGG, PCA, SVM |

## 3) Methods

Considering the challenges involved in the recognition of Persian handwritings, which were expiated throughout the previous sections, in this study, the DenseNet architecture is utilized. As their main advantage in comparison to the rest of the existing state-of-the-art alternatives, while still being deep enough to capture the distinctive nuances based on an efficient selection of features, they have significantly less parameters, which makes them more suitable, in terms of avoiding overfitting, for devising an OCR system on the basis of the existing relatively small databases of handwritten Persian text. They usually have 60,000 to 70,000 parameters, being significantly smaller than the number of parameters of ResNet architects, i.e. about 25 to 26 million. Their structure is inspired by the design of both ResNet [25] and Inception [26] architectures, where the blocks are essentially a sequence of layers whose input is a set of all the previous feature maps in an appended form.

A network of the above type with $L$ layers has $L$ connections, where each densblock has $l(l+1)/2$ direct connections. Each layer is directly connected to all others in a feed-forward fashion, and the feature map of each of the preceding layers is considered as a separate input. Multiple inputs $H_l()$, each of which consists of three manual normalization operators, ReLU and a 3×3 convolution, are combined as a single tensor for pre-activation. The transition layers are the ones mediating between the densblocks, which are composed of a batch normalization unit and a 6×6 convolution layer, being then followed by a 2×2 average pooling module with stride 2. Compared to the rest of state-of-the-art architectures, DenseNet requires less computation to achieve high performance. In this study, for recognizing digits and letters, four different versions of DenseNet architectures, namely, DenseNet121, DenseNet161, DenseNet169 and DenseNet201, have been utilized. As shown in Fig. 2(a), the numbers 121,161,169 and 201 stand for the number of layers with trainable weights. It should be noted that for recognizing words, the last layers of the aforementioned networks have been changed such that the fully connected and dropout layers are repeated twice, followed by Softmax layers, resulting in a new architecture called *DenseNet+* being proposed in this paper, which is illustrated in Fig. 2(b). Moreover, the Xception model [27] has been tested as well, for the sake of comparison. The details of the experiments will be presented in the following sections.

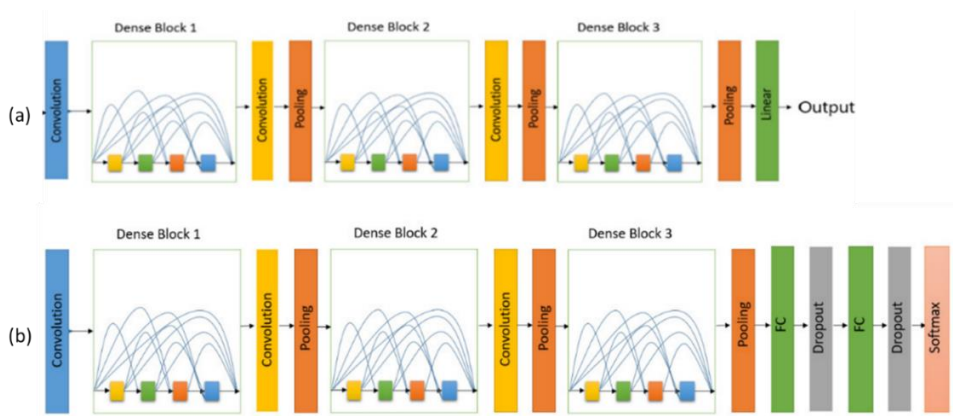

*Figure 2. Schematic illustration of the DenseNet architecture: (a) Used for digits and characters, (b) Used for words.*

## 3-1) Databases

In this study, the Sadri and HODA databases are utilized for evaluating the proposed method. The Sadri database contains pieces of text, words, numbers, dates consisting of numbers and words, letters, signs and symbols. The number of words in the Sadri database is 70,000, collecting which has been performed in a random manner by 500 Persian-language authors, of which 250 were men and 250 women, 10% of whom were left-handed. Each digit has been written 20 times by each author, resulting in 97194 images of the 10 digits, i.e. about 9,000 for each. Moreover, images of all the 32 Persian letters have been collected based on the different forms they may take depending on their place within a word. In total, there are 42,962 images of letters for a total of 86 classes, each being represented by about 500. Example images of digits, letters and words from this database are shown in Fig. 3. Similarly, the HODA database, as illustrated with a few samples in Fig. 4, is composed of digits and letters written by high school and B.Sc. students in Iran, which have been scanned at a resolution of 200 Dots Per Inch (DPI). When it comes to digits, there are a total of 60,000 and 20,000 images for training and test procedures, respectively. There are also 70,645 and 17,706 images of letters for training and test purposes, respectively.

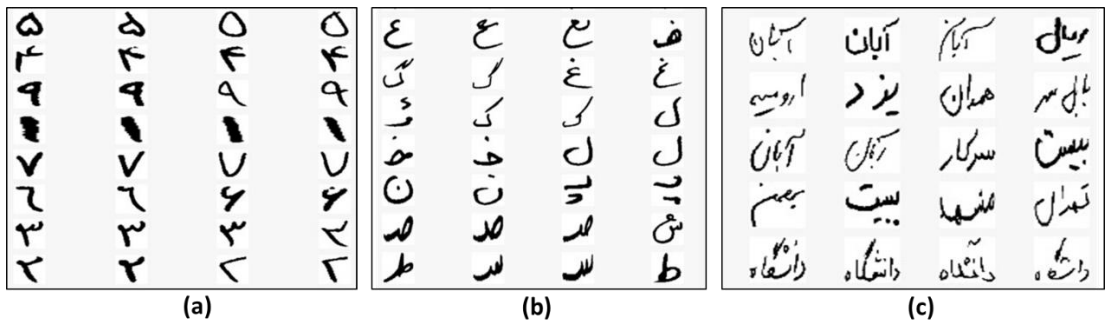

Figure 3. Sample images from the Sadri database: (a) Digits, (b) Letters, (c) Words.

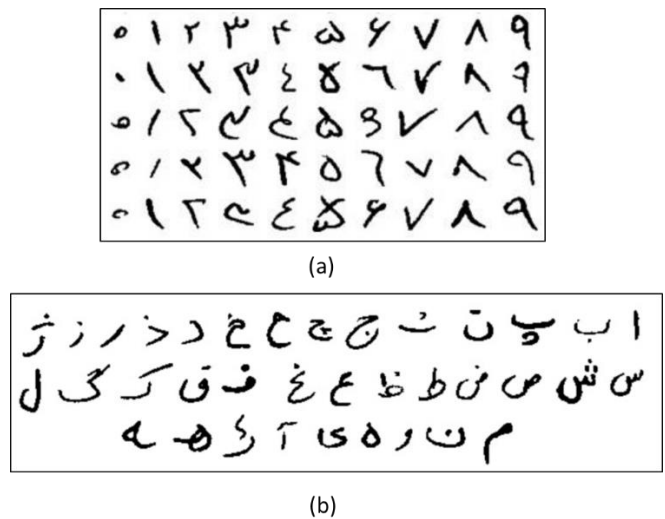

Figure 4.  Sample images from the HODA database: (a) Digits, (b) Letters.

3-2) Preprocessing

Although resizing the images to the same dimensions is beneficial in terms of consistency and simplicity, enlarging or shrinking them excessively may cause blurriness or significant data loss, respectively, which may affect the end result. Moreover, enlarging the images has a considerable computational cost, which grows significantly with the target dimensions. Therefore, accounting for the differences in the dimensions of the raw image, the relevant averages and standard deviations have been calculated, resulting in a choice of 80×80 and 64×64 pixels as the resizing dimensions for the images from the Sadri and HODA databases, respectively, which was achieved using the inter-area interpolation functionality of the OpenCV library [28]. In addition, morphological operation such as dilation for extending the bright areas and median filter for noise reduction and edge smoothing were employed. Fig. 5 depicts examples of the results of applying the foregoing operations on raw images for illustration. Moreover, in order to alleviate the lack of training samples, data augmentation and Test Time Augmentation (TTA) were applied to the training and test data, respectively. More specifically, for characters and digits, operations such as rotation, median and Gaussian filtering and shift, and for words, shift-scale, rotation and horizontal flipping were resorted to.

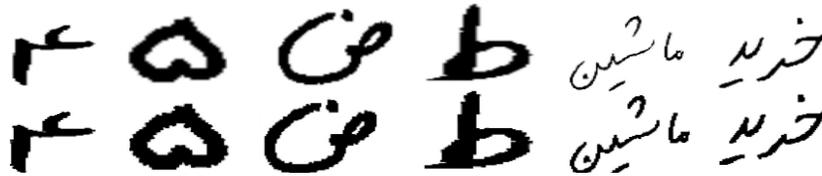

*Figure 5. Sample images and their preprocessed counterparts shown in the top and bottom rows, respectively.*

3-3) Training

In the first round of the experiments carried out for the purpose of this paper, each database is divided into training, validation and test sets, as detailed in Table 2, being followed by applying the DenseNet121, DenseNet161 and DenseNet169 architectures, and then performing augmentation on the one which resulted in the highest recognition rate.

When it comes to the HODA database, the images of digits are divided into subsets of 60%, 20% and 20% of the total number of samples for training, validation and test, being 70%, 20% and 10% in the case of letters, respectively. The number of epochs is 150 with a batch size of 2048, a learning rate of 0.09 and a Stochastic Gradient Descent (SGD) [29] optimizer, as well as categorical cross entropy as the loss function, a regularization parameter of 0.0001 and a drop out of 0.15 before the fully connected layer were used. The recognition rate resulted from each architecture is shown in Fig. 6, which demonstrates the superior performance of the DenseNet121 architecture on both digits and letters, being even further improved to 99.49% and 98.1%, respectively, through augmentation.

Similarly, the Sadri database is divided into subsets of 60%, 20% and 20% of the total number of digit samples for training, validation and test, being 80%, 15% and 5% in the case of letters, as well as 73%, 12% and 15% in the case of words, respectively. The DesneNet121, DenseNet161 and DenseNet169 are applied, while this time, the drop out is set to 0.1, and a batch size of 1024 is chosen for letters. The Xception and DenseNet201 architectures are also applied to the words so as to achieve an ensemble structure taking advantage of majority voting. In the foregoing scenario, majority voting is exercised in order to make the recognition decision as a consensus among the inferences made based on the different variants of each sample image. As shown in Fig. 7, 270 epochs are considered with a batch size of 1024 and a learning rate of 0.0008, as well as Adam [30] and categorical cross entropy as the optimizers. The resulting recognition rates are shown in Fig. 8.

In the next round, *k*-fold cross-validation [31] is used, in which *k* is 10, where the database is divided into roughly equal-sized subsets, each being used exactly once as the test set, with the remaining data being utilized for training. The DenseNet121 architecture is employed, along with an alternative version taking advantage of randomly applying TTA and majority voting as well, being referred to as DenseNet121-TTA for brevity. The resulting recognition rates are listed in Table 3.

Table 2. The number and percentage of the samples utilized for training, validation and test from each database, along with the relevant file types, as well as the raw and preprocessed dimensions, for each of the digit, letter and word categories.

| Database | Training | Validation | Test | Raw dimensions | Preprocessed dimensions |
|---|---|---|---|---|---|
| HODA (letters) | 63580 / 70% | 17706 / 20% | 7065 / 10% | 22×21 to 37×36 | 64×64 |
| HODA (digits) | 60000 / 60% | 22352 / 20% | 20000 / 20% | 21×14 to 35×26 | 64×64 |
| Sadri (words) | 51312 / 73% | 8139 / 12% | 10496 / 15% | 21×22 to 221×440 | 80×80 |
| Sadri (letters) | 34369 / 80% | 6445 / 15% | 2148 / 5% | 41×37 to 73×77 | 64×64 |
| Sadri (digits) | 60000 / 60% | 17194 / 20% | 20000 / 20% | 28×24 to 52×44 | 64×64 |

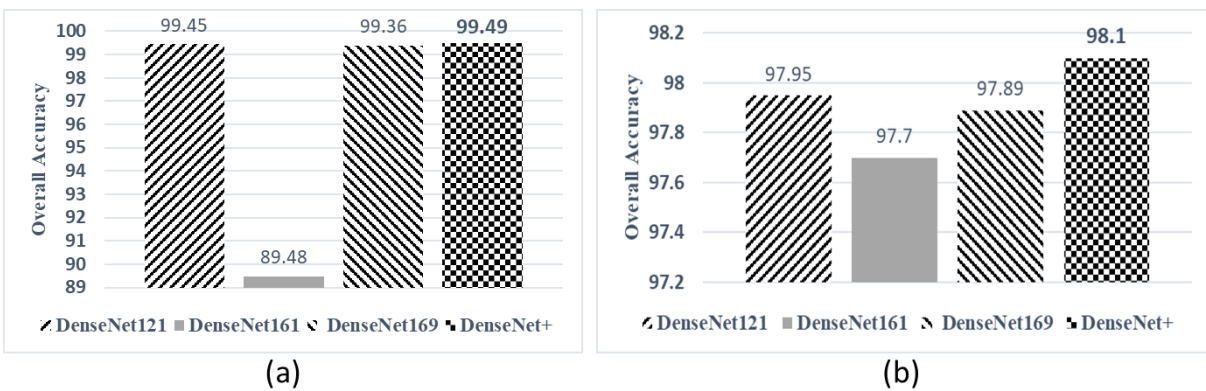

Figure 6. The OCR rates on the HODA database using different DenseNet architectures, as well as the proposed DenseNet+, on (a) Digits, (b) Characters.

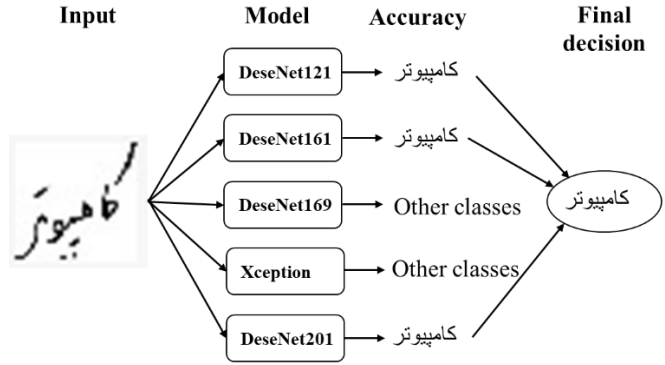

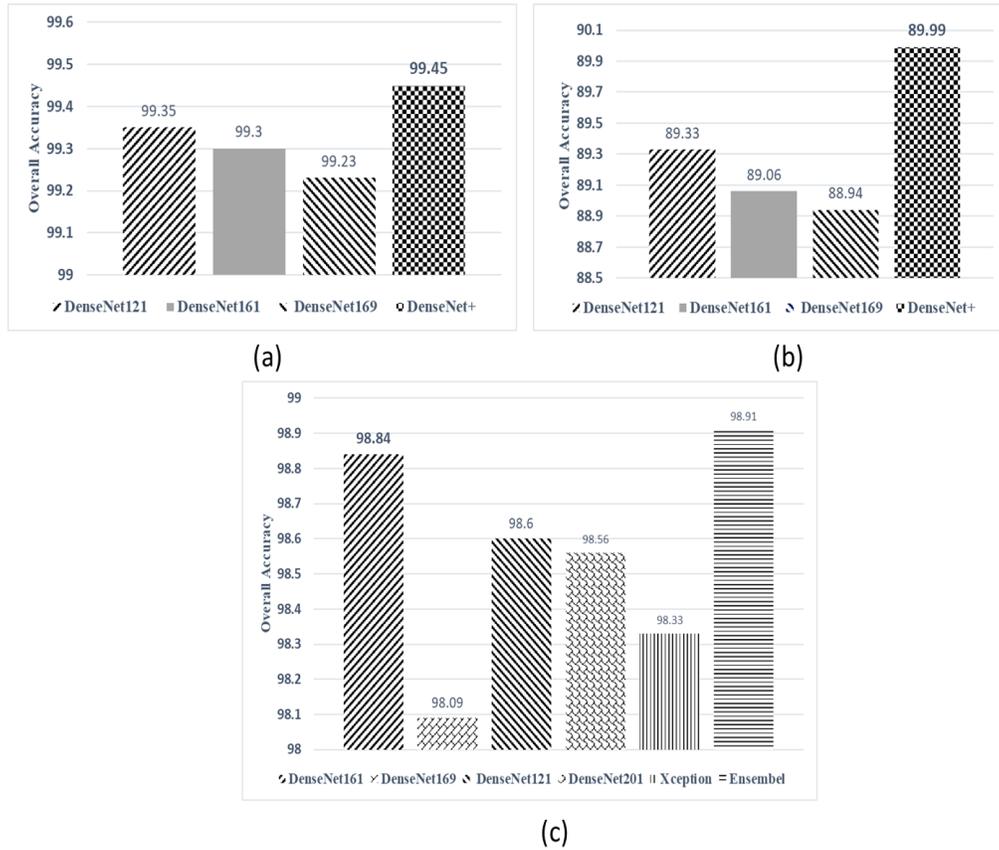

Figure 7. Schematic illustration of the ensemble structure using majority voting for word recognition.

Figure 8. The OCR rates on the Sadri database using different DenseNet architectures, as well as the proposed DenseNet+, on (a) Digits, (b) Characters and (c) Words, where an ensemble structure has also been tested on the latter.

Table 3. The percentage of correctly recognized digits, letters and words from the HODA and Sadri databases using 10-fold cross-validation and different DenseNet structures.

| Database | Data | Architecture | 1 | 2 | 3 | 4 | 5 | 6 | 7 | 8 | 9 | 10 | Mean |
|---|---|---|---|---|---|---|---|---|---|---|---|---|---|
| HODA | Digits | DenseNet121 | 99.79 | 99.66 | 99.74 | 99.58 | 99.74 | 99.81 | 99.70 | 99.75 | 99.76 | 99.71 | **99.72** |
| | | DenseNet121+TTA | 99.79 | 99.69 | 99.74 | 99.55 | 99.71 | 99.80 | 99.70 | 99.74 | 99.78 | 99.71 | **99.72** |
| | Letters | DenseNet121 | 98.38 | 98.23 | 98.50 | 98.29 | 98.30 | 98.24 | 98.04 | 98.15 | 98.13 | 98.16 | 98.24 |
| | | DenseNet121+TTA | 98.46 | 98.36 | 98.55 | 98.31 | 98.42 | 98.45 | 98.17 | 98.11 | 98.14 | 98.24 | **98.32** |

| | | | | | | | | | | | | |
|---|---|---|---|---|---|---|---|---|---|---|---|---|
| **Sadri** | Digits | DenseNet121 | 99.46 | 99.54 | 99.37 | 99.46 | 99.52 | 99.45 | 99.48 | 99.46 | 99.39 | 99.38 | 99.45 |
| | | DenseNet121+TTA | 98.49 | 99.53 | 99.39 | 99.54 | 99.58 | 99.44 | 99.53 | 99.50 | 99.48 | 99.41 | **99.49** |
| | Letters | DenseNet121 | 89.23 | 89.72 | 88.91 | 89.74 | 90.16 | 89.20 | 89.71 | 89.71 | 89.89 | 89.44 | 89.57 |
| | | DenseNet121+TTA | 89.61 | 90.07 | 89.51 | 90.16 | 90.42 | 89.65 | 89.83 | 90.38 | 90.47 | 89.58 | **89.97** |
| | Words | DenseNet121 | 98.54 | 98.90 | 98.90 | 98.79 | 98.71 | 98.60 | 98.50 | 98.64 | 98.73 | 98.71 | 98.70 |
| | | DenseNet161 | 98.69 | 98.86 | 98.64 | 98.96 | 98.87 | 98.73 | 98.71 | 98.76 | 98.78 | 98.87 | **98.82** |

## 4) Results and Discussion

In this section, we compare the performance of the proposed method with the existing state-of-the-art alternatives based on the recognition rates listed in Table 4 for digits, letters and words, where the entries corresponding to the proposed method stand for the best results obtained using the various architectures described throughout the preceding sections. For the digits from the HODA database, our 10-fold cross validation, which resulted in a recognition rate of 99.72%, outperforms the best systems introduced in the literature heretofore. Similarly, on the Sadri dataset, we achieved a recognition rate of 99.72%, while that of Sadri et al. [6] has been 98.57%. Regarding letters, due to the limited number of prior work, we have also provided the result of one of them on the TMU database, where it could be clearly observed that the proposed method improves the current state-of-the-art on the HODA database, i.e. the recognition rate of 97.7% reported by Sarvaramini et al. [21] to 98.32% and 98.10%, using different architectures. Moreover, the proposed method has been tested on the letters from the Sadri database, which resulted in desirable recognition rates of 89.99% and 89.97%. Last but not least, Studies on handwritten Persian words are extremely rare, were our study of the words from the Sadri database is, to the authors' knowledge, the first one reported in the relevant literature. Again, it can be seen that the recognition rates of 98.82% and 98.70% achieved in this paper surpass the ones obtained in the previous studies.

Table 4. Comparison of our method with the state-of-the-art alternatives in terms of correctly recognizing handwritten Persian digits, letters and words, based on different databases.

| Data type | Database | Reference | Recognition rate |
|---|---|---|---|
| Digits | HODA | Alaei et al. [32] | 99.37% |
| | | Hosseini et al. [33] | 97.12% |
| | | Parseh et al. [11] | 99.07% |
| | | Montazer et al. [16] | 99.30% |
| | | Al-wajih et al. [34] | 99.13% |
| | | Sajedi et al. [15] | 99.07% |
| | | Khorashadizadeh et al. [4] | 99.58% |
| | | Bossaghzadeh et al. [24] | 99.69% |
| | | Nanehkaran et al. [18] | 99.45% |
| | | Safarzadeh et al. [35] | 99.37% |
| | | Parseh et al. [17] | 99.56% |
| | | Akhlaghi et al. [20] | 99.34% |
| | | Our method | 99.49% |
| | | **Our method (10-fold)** | **99.72%** |
| | Sadri | Sadri et al. [10] | 98.57% |
| | | **Our method** | **99.72%** |
| | | **Our method (10-fold)** | **99.72%** |
| Characters | HODA | Hajihashemi et al. [12] | 88.9 % |

|  |  | Sarvaramini et al. [21] | 97.7% |
|  |  | Roohi et al. [22] | 97.1% |
|  |  | **Our method** | **98.10%** |
|  |  | **Our method (10-fold)** | **98.32%** |
|  | TMU | Ghods et al. [13] | 95.9% |
|  | Sadri | **Our method** | **89.99%** |
|  |  | **Our method (10-fold)** | **89.97%** |
|  | Iranshahr | Arani et al. [14] | 97.93% |
|  |  | Ghadikolaie et al. [19] | 90.5% |
| **Words** | Sadri | **Our method** | **98.82%** |
|  |  | **Our method (10-fold)** | **98.70%** |

For illustration, confusion matrices representing the performance of the proposed method on the recognition of handwritten Persian digits are shown in Fig. 9 for the HODA and Sadri databases, where the numbers on the diagonals stand for the count of the correctly recognized samples from the corresponding classes. They indicate the fact that the model's error is not biased toward any of the specific classes, but is rather distributed to all of the classes. It is also observable that some of the digits that, as shown in Fig. 1(a), are very similar in appearance, e.g. "3" and "2", are more likely to be misclassified. It should be noted that the confusion matrices representing the performance of the proposed method on letters and words are also accessible at the following webpage: .

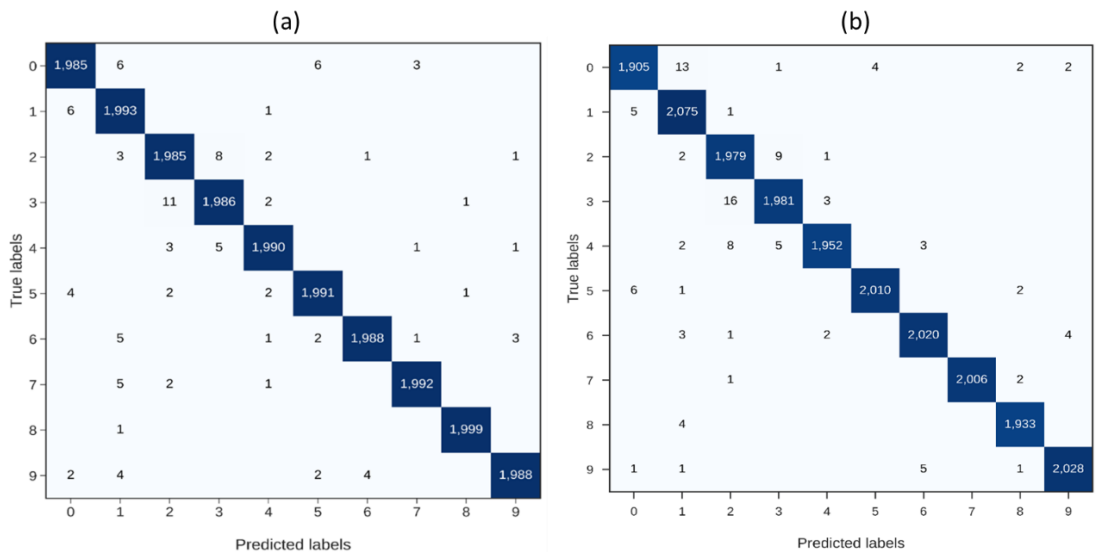

*Figure 7. Confusion matrices representing the performance of the proposed method on the recognition of handwritten Persian digits: (a) The HODA database, (b) The Sadri database.*

## 5) Conclusion

In this paper, a method was proposed to recognize handwritten Persian digits, letters and words. Various DenseNet architectures were employed as fully convolutional deep neural networks, due to their known capability of extracting efficient image-based features. Specifically, the Densenet121, DenseNet161 and DenseNet169 architectures were utilized, while for recognizing words, Xception and DenseNet201 were tested as well. Training of the networks was achieved both through dividing the databases into training, validation and test sets and k-fold cross validation. Two comprehensive Persian handwriting databases, namely HODA and Sadri, were considered for evaluating the proposed method through comparing it with the state-of-the-art methods having led to the highest optical character recognition rates reported in the literature heretofore, clearly revealing the superiority of the proposed method. On the HODA database, the state-of-the-art recognition rates were improved to 99.72% and 89.99% for digits and characters, and on the Sadri database, to 99.72%, 98.32% and 98.82% for digits, characters and words, respectively. Nevertheless, extending the system for other languages and evaluating its robustness against numerous linguistic and paralinguistic variations remains a challenge which needs to be investigated and tackled as a possible research direction in the relevant future studies.

## 6) Acknowledgment

This research has been financed by the European Social Fund via IT Academy programme and by the Estonian Research Council via grant COVSG24.

## 7) Disclosures

The authors declare no conflict of interest.

## 8) References

[1] S. Pansare and D. Joshi, "A Survey on Optical Character Recognition Techniques," *International Journal of Science and Research (IJSR),* vol. 3, no. 12, pp. 1247-1249, 2014.
[2] E. Borovikov, "A survey of modern optical character recognition techniques," *arXiv preprint arXiv:1412.4183,* 2014.
[3] D. ping Tian, "A review on image feature extraction and representation techniques," *International Journal of Multimedia and Ubiquitous Engineering,* vol. 8, no. 4, pp. 385-396, 2013.
[4] S. Khorashadizadeh and A. Latif, "Arabic/Farsi Handwritten Digit Recognition usin Histogra of Oriented Gradient and Chain Code Histogram," *International Arab Journal of Information Technology (IAJIT),* vol. 13, no. 4, 2016.
[5] M. Dehghan and K. Faez, "Farsi handwritten character recognition with moment invariants," in *Proceedings of 13th International Conference on Digital Signal Processing*, 1997, vol. 2, pp. 507-510: IEEE.
[6] X.-X. Niu and C. Y. Suen, "A novel hybrid CNN–SVM classifier for recognizing handwritten digits," *Pattern Recognition,* vol. 45, no. 4, pp. 1318-1325, 2012.
[7] G. E. Hinton and R. R. Salakhutdinov, "Reducing the dimensionality of data with neural networks," *science,* vol. 313, no. 5786, pp. 504-507, 2006.
[8] G. Huang, Z. Liu, L. Van Der Maaten, and K. Q. Weinberger, "Densely connected convolutional networks," in *Proceedings of the IEEE conference on computer vision and pattern recognition*, 2017, pp. 4700-4708.
[9] H. Khosravi and E. Kabir, "Introducing a very large dataset of handwritten Farsi digits and a study on their varieties," *Pattern recognition letters,* vol. 28, no. 10, pp. 1133-1141, 2007.
[10] J. Sadri, M. R. Yeganehzad, and J. Saghi, "A novel comprehensive database for offline Persian handwriting recognition," *Pattern Recognition,* vol. 60, pp. 378-393, 2016.

[11] M. J. Parseh and M. Meftahi, "A new combined feature extraction method for Persian handwritten digit recognition," *International Journal of Image and Graphics,* vol. 17, no. 02, p. 1750012, 2017.

[12] V. Hajihashemi, M. M. A. Ameri, A. A. Gharahbagh, and A. Bastanfard, "A pattern recognition based Holographic Graph Neuron for Persian alphabet recognition," in *2020 International Conference on Machine Vision and Image Processing (MVIP)*, 2020, pp. 1-6: IEEE.

[13] V. Ghods and M. K. Sohrabi, "Online Farsi Handwritten Character Recognition Using Hidden Markov Model," *JCP,* vol. 11, no. 2, pp. 169-175, 2016.

[14] S. A. A. A. Arani, E. Kabir, and R. Ebrahimpour, "Combining RtL and LtR HMMs to recognise handwritten Farsi words of small-and medium-sized vocabularies," *IET Computer Vision,* vol. 12, no. 6, pp. 925-932, 2018.

[15] H. Sajedi, "Handwriting recognition of digits, signs, and numerical strings in Persian," *Computers & Electrical Engineering,* vol. 49, pp. 52-65, 2016.

[16] G. A. Montazer, M. A. Soltanshahi, and D. Giveki, "Farsi/Arabic handwritten digit recognition using quantum neural networks and bag of visual words method," *Optical Memory and Neural Networks,* vol. 26, no. 2, pp. 117-128, 2017.

[17] M. Parseh, M. Rahmanimanesh, and P. Keshavarzi, "Persian Handwritten Digit Recognition Using Combination of Convolutional Neural Network and Support Vector Machine Methods," *INTERNATIONAL ARAB JOURNAL OF INFORMATION TECHNOLOGY,* vol. 17, no. 4, pp. 572-578, 2020.

[18] Y. Nanehkaran, D. Zhang, S. Salimi, J. Chen, Y. Tian, and N. Al-Nabhan, "Analysis and comparison of machine learning classifiers and deep neural networks techniques for recognition of Farsi handwritten digits," *The Journal of Supercomputing,* pp. 1-30, 2020.

[19] M. F. Y. Ghadikolaie, E. Kabir, and F. Razzazi, "Sub-word Based Offline Handwritten Farsi Word Recognition Using Recurrent Neural Network," *ETRI Journal,* vol. 38, no. 4, pp. 703-713, 2016.

[20] M. Akhlaghi and V. Ghods, "Farsi handwritten phone number recognition using deep learning," *SN Applied Sciences,* vol. 2, no. 3, pp. 1-10, 2020.

[21] F. Sarvaramini, A. Nasrollahzadeh, and M. Soryani, "Persian handwritten character recognition using convolutional neural network," in *Electrical Engineering (ICEE), Iranian Conference on*, 2018, pp. 1676-1680: IEEE.

[22] B. Alizadehashraf and S. Roohi, "Persian handwritten character recognition using convolutional neural network," in *2017 10th Iranian Conference on Machine Vision and Image Processing (MVIP)*, 2017, pp. 247-251: IEEE.

[23] E. Farahbakhsh, E. Kozegar, and M. Soryani, "Improving persian digit recognition by combining data augmentation and AlexNet," in *2017 10th Iranian Conference on Machine Vision and Image Processing (MVIP)*, 2017, pp. 265-270: IEEE.

[24] A. Bossaghzadeh, "Improving Persian Digit Recognition by Combining Deep Neural Networks and SVM and Using PCA," in *2020 International Conference on Machine Vision and Image Processing (MVIP)*, 2020, pp. 1-5: IEEE.

[25] K. He, X. Zhang, S. Ren, and J. Sun, "Deep residual learning for image recognition," in *Proceedings of the IEEE conference on computer vision and pattern recognition*, 2016, pp. 770-778.

[26] C. Szegedy *et al.*, "Going deeper with convolutions," in *Proceedings of the IEEE conference on computer vision and pattern recognition*, 2015, pp. 1-9.

[27] F. Chollet, "Xception: Deep learning with depthwise separable convolutions," in *Proceedings of the IEEE conference on computer vision and pattern recognition*, 2017, pp. 1251-1258.

[28]  W. L. Smith *et al.*, "Prostate volume contouring: a 3D analysis of segmentation using 3DTRUS, CT, and MR," *International Journal of Radiation Oncology* Biology* Physics,* vol. 67, no. 4, pp. 1238-1247, 2007.
[29]  L. Bottou, "Stochastic gradient descent tricks," in *Neural networks: Tricks of the trade*: Springer, 2012, pp. 421-436.
[30]  D. P. Kingma and J. Ba, "Adam: A method for stochastic optimization," *arXiv preprint arXiv:1412.6980,* 2014.
[31]  T. Fushiki, "Estimation of prediction error by using K-fold cross-validation," *Statistics and Computing,* vol. 21, no. 2, pp. 137-146, 2011.
[32]  A. Alaei, U. Pal, and P. Nagabhushan, "Using modified contour features and SVM based classifier for the recognition of Persian/Arabic handwritten numerals," in *2009 Seventh International Conference on Advances in Pattern Recognition*, 2009, pp. 391-394: IEEE.
[33]  M. S. Hosseini-Pozveh, M. Safayani, and A. Mirzaei, "Interval Type-2 Fuzzy Restricted Boltzmann Machine," *IEEE Transactions on Fuzzy Systems,* 2020.
[34]  E. Al-wajih and R. Ghazali, "Improving the Accuracy for Offline Arabic Digit Recognition Using Sliding Window Approach," *Iranian Journal of Science and Technology, Transactions of Electrical Engineering,* pp. 1-12, 2020.
[35]  V. M. Safarzadeh and P. Jafarzadeh, "Offline Persian Handwriting Recognition with CNN and RNN-CTC," in *2020 25th International Computer Conference, Computer Society of Iran (CSICC)*, 2020, pp. 1-10: IEEE.